\documentclass[11pt,a4paper]{article}
\usepackage{paper-layout}
\usepackage{multirow}
\usepackage{colortbl}
\usepackage{needspace}
\usepackage{wrapfig}
\SetSymbolFont{operators}{normal}{OT1}{cmr}{m}{n}
\SetSymbolFont{operators}{bold}{OT1}{cmr}{bx}{n}
\SetMathAlphabet{\mathit}{normal}{OT1}{cmr}{m}{it}
\SetMathAlphabet{\mathit}{bold}{OT1}{cmr}{bx}{it}
\definecolor{AblDrop}{HTML}{426A80}
\definecolor{AblRise}{HTML}{5586A0}
\definecolor{AblFull}{HTML}{EDF5F8}
\definecolor{MainBest}{HTML}{BCD7E8}
\definecolor{MainSecond}{HTML}{E6F0F7}
\newcommand{\abldown}[1]{\textcolor{AblDrop}{\ensuremath{_{\scriptstyle\downarrow\,#1}}}}
\newcommand{\ablup}[1]{\textcolor{AblRise}{\ensuremath{_{\scriptstyle\uparrow\,#1}}}}
\newcommand{\ablsame}{\textcolor{AblDrop}{\ensuremath{_{\scriptstyle\,0.0}}}}
\newcommand{\method}{\textsc{FrontierSD}}
\newcommand{\sg}{\operatorname{sg}}

\newcommand{\PlaceFirstPageHeader}{\AddToShipoutPictureFG*{%
  \put(\LenToUnit{\dimexpr1in+\oddsidemargin\relax},\LenToUnit{\dimexpr\paperheight-2.35cm\relax}){%
    \makebox(0,0)[l]{\makebox[\textwidth][l]{%
      \raisebox{-0.5\height}{\includegraphics[trim={0bp 16bp 21bp 7bp},clip,height=0.82cm]{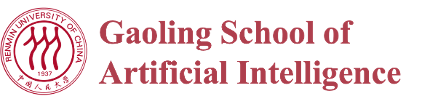}}%
      \hspace{0.75cm}%
      \raisebox{-0.5\height}{\includegraphics[trim={0bp 0bp 1bp 0bp},clip,height=0.82cm]{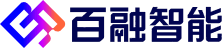}}\hfill}}}%
  \put(\LenToUnit{\dimexpr1in+\oddsidemargin\relax},\LenToUnit{\dimexpr\paperheight-3.12cm\relax}){%
    \makebox(0,0)[lt]{\rule{\textwidth}{0.4pt}}}%
}}
\newcommand{\projectlink}{\url{https://namespace-eri.github.io/FrontierSD/}}
\newcommand{\ghlink}{\url{https://github.com/namespace-ERI/FrontierSD}}
\newcommand{\hfDataLink}{\href{https://huggingface.co/datasets/namespace-ERI/frontiersd-data}{\nolinkurl{namespace-ERI/frontiersd-data}}}
\newcommand{\hfModelLink}{\href{https://huggingface.co/namespace-ERI/frontiersd-model}{\nolinkurl{namespace-ERI/frontiersd-model}}}
\newcommand{\projectlinks}{\par\vspace{0.35em}\begingroup
  \raggedright\fontsize{8pt}{9.6pt}\selectfont
  \renewcommand{\arraystretch}{1.08}\setlength{\tabcolsep}{0pt}
  \begin{tabular}{@{}l@{\hspace{0.55em}}>{\bfseries\color{themecolor}}l@{\hspace{1.1em}}l@{}}
  \raisebox{-1.1pt}{\includegraphics[height=7.8pt]{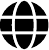}} & Project Page & \projectlink \\
  \raisebox{-1.1pt}{\includegraphics[height=7.8pt]{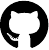}} & GitHub & \ghlink \\
  \raisebox{-1.1pt}{\includegraphics[height=7.8pt]{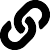}} & Hugging Face & \hfModelLink\hspace{2em}\hfDataLink
  \end{tabular}\par\endgroup}

\posttitle{\par\endgroup\vspace{0.2em}}
\postauthor{\par\endgroup\vspace{0.1em}}
\title{{\fontsize{18pt}{22pt}\selectfont\bfseries
\textcolor{themecolor}{Pushing the Error Frontier:}\\[2pt]
Recursive Self-Improvement of Long-Horizon Agents\\
via Frontier-Aligned Self-Distillation}}
\author{%
Yuyang Hu\textsuperscript{1}, Yashu Liu\textsuperscript{2}, Tong Zhao\textsuperscript{1}, Xuanyu Zhou\textsuperscript{1},\\
Hongjin Qian\textsuperscript{3}, Jiajie Jin\textsuperscript{1}, Ling Huang\textsuperscript{2}, Zhicheng Dou\textsuperscript{1,\textdagger}\\[5pt]
{\small\textsuperscript{1}Gaoling School of Artificial Intelligence, Renmin University of China}\\
{\small\textsuperscript{2}Bairong AI Inc.\qquad\textsuperscript{3}Independent Researcher}\\[3pt]
{\footnotesize\texttt{\{yuyang.hu, dou\}@ruc.edu.cn}}\\[3pt]
{\footnotesize\textsuperscript{\textdagger}Corresponding author}
}

\date{}
\hypersetup{
 pdftitle={Pushing the Error Frontier: Recursive Self-Improvement of Long-Horizon Agents via Frontier-Aligned Self-Distillation},
 pdfauthor={Yuyang Hu, Yashu Liu, Tong Zhao, Xuanyu Zhou, Hongjin Qian, Jiajie Jin, Ling Huang, Zhicheng Dou},
 pdfsubject={FrontierSD},pdfkeywords={}}
\begin{document}
\PlaceFirstPageHeader
\maketitle
\vspace{-1.5em}
\begin{abstract}
LLM-based agents can now handle longer tasks, such as web shopping and browser navigation. They often do well in the early steps but get stuck later. Outcome-based training such as group relative policy optimization (GRPO) offers little feedback about where these bottlenecks lie or how to overcome them. Another recent approach, on-policy distillation (OPD), provides dense feedback, but dense feedback alone does not specify where teaching should begin or when a rule should stop supervising. We argue that teaching should focus on the current bottleneck and advance as the agent learns. We introduce the error frontier: the position separating a mastered prefix, which the agent executes reliably, from an unmastered suffix, where progress becomes less reliable and success less likely. Our insight is to turn hindsight at this position into supervision for overcoming the bottleneck. We propose \method{} to combine frontier localization with local self-distillation. Entropy recall and action-support comparison select the teaching position, where the agent extracts a hindsight rule and distills rule-guided predictions in the original decision context. Rule influence controls teaching strength and scope, alongside full-trajectory GRPO. After initialization with externally labeled examples, the same agent generates fresh trajectories and rules for successive updates, forming a recursive self-improvement loop. We evaluate Qwen3-1.7B and Qwen3-4B in four environments without guidance at test time. FrontierSD improves success on WebShop, ALFWorld seen, and MiniHack at both scales. The 4B gains over the strongest reported baselines are 4.3 and 10.8 percentage points on WebShop and ALFWorld seen, respectively; success on ALFWorld unseen reaches 91.8\%. During training, teaching generally moves to later positions and guidance is triggered less often, consistent with an expanding mastered prefix.
\projectlinks
\end{abstract}
\vspace{0.50cm}

\noindent\begin{minipage}{\linewidth}
    \centering
    \includegraphics[trim={21.328pt 68.781pt 13.552pt 62.739pt},clip,width=\textwidth]{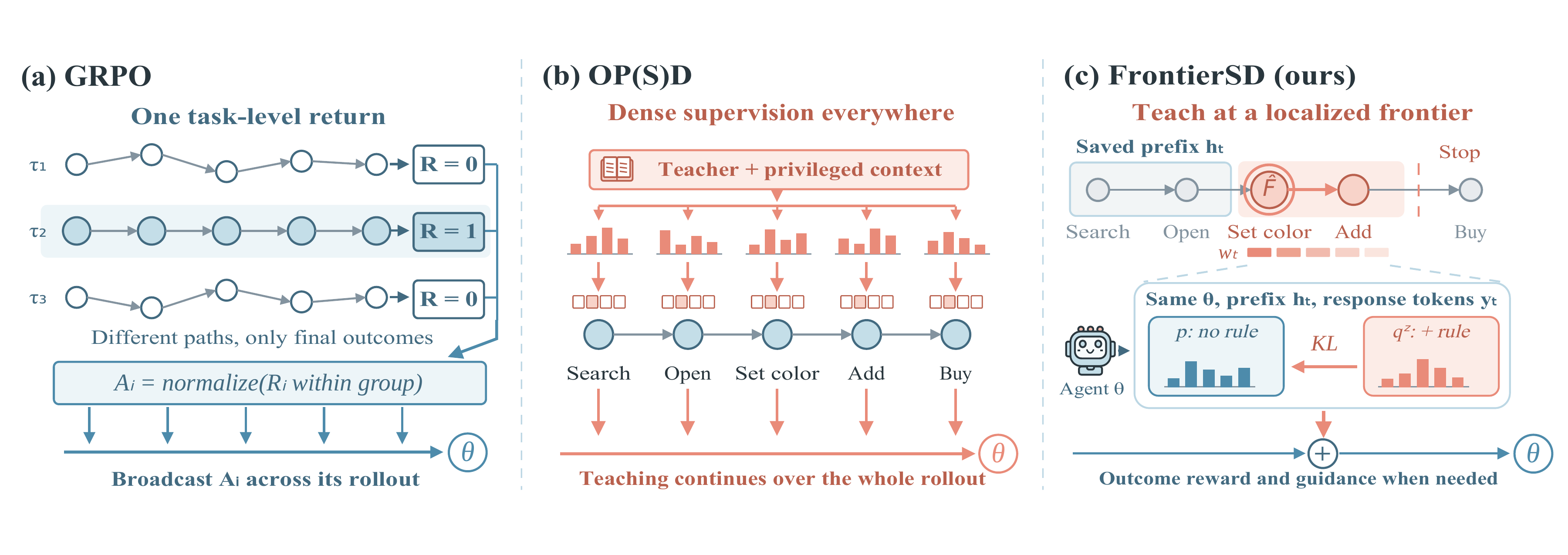}
    \captionof{figure}{Training feedback: (a) task-outcome rewards; (b) trajectory-wide distillation; (c) frontier-local OPD with full-trajectory GRPO. A rule is a short instruction extracted from experience, such as checking the requested color. The inset scores the same response prefix with an unguided student and a stop-gradient guided teacher.}
    \label{fig:intuition}
\end{minipage}
\clearpage

\section{Introduction}

\begin{wrapfigure}{r}{0.51\textwidth}
\vspace{-6pt}
\centering
\captionsetup{font=papersubcaption}
\includegraphics[width=\linewidth]{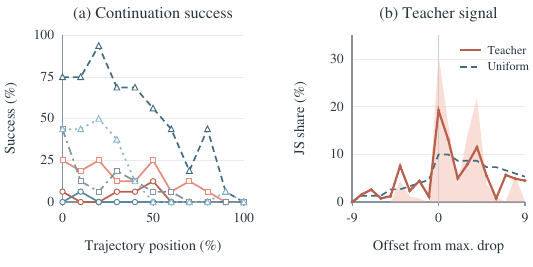}
\caption{ALFWorld diagnostics with base Qwen3-1.7B (15 tasks). (a) Each curve is one failed trajectory; success is the fraction of 16 continuations that finish from a replayed prefix within a 50-action total budget. (b) Normalized action-token Jensen--Shannon divergence under a successful-trajectory reference, aligned to the largest success-drop interval. Shading: interquartile range. Protocol: Appendix~\ref{app:intro-observations}.}
\label{fig:intro-observations}
\end{wrapfigure}

LLM-based agents can complete tasks requiring increasingly many steps, including web shopping and browser navigation~\citep{yao2023webshopscalablerealworldweb,zhou2024webarenarealisticwebenvironment,drouin2024workarenacapablewebagents,cai2026sageflash}. These tasks require coordinated reasoning, action, and planning in response to environment feedback~\citep{yao2023reactsynergizingreasoningacting,zhou2024languageagenttreesearch,hu2026compassmem,hu2025agentmemory}. Yet sustaining correct decisions and steady progress over many turns remains difficult~\citep{liu2025agentbenchevaluatingllmsagents,wang2025ragenunderstandingselfevolutionllm,hu2026sam}.

An early mistake changes what the agent observes later, affecting subsequent progress~\citep{ross2011reductionimitationlearningstructured}. The final reward reports the outcome but gives little indication of where the agent went off track~\citep{xi2025agentprmprocessrewardmodels,wang2025ragenunderstandingselfevolutionllm}. For example, an agent may search and browse products successfully, then go off track when selecting options or checking attributes. Our intuition resembles a level-based game: a player must clear the current stage to reach later ones. Teaching should likewise help the agent through the stage currently blocking its progress.

To examine this pattern, we let base Qwen3-1.7B continue from different positions in failed ALFWorld trajectories (Figure~\ref{fig:intro-observations}(a)). Completion probability varies along the trajectory, and the stages where it drops differ across trajectories. This suggests that teaching should begin at the difficulty the agent encounters in its current trajectory.

Existing training approaches have their own limitations in addressing this problem. Supervised fine-tuning teaches agents from expert trajectories, but these examples are costly to collect, and their prefixes may differ from the contexts the learner actually visits~\citep{ross2011reductionimitationlearningstructured,agarwal_-policy_2024}. Outcome-based RL learns from the agent's own experience~\citep{wang2025ragenunderstandingselfevolutionllm}, while reflective agents reuse feedback in subsequent attempts~\citep{shinn2023reflexionlanguageagentsverbal,jin2026arbor,qian2026arex2}. Yet task-level feedback leaves intermediate credit assignment unresolved~\citep{xi2025agentprmprocessrewardmodels,lu_self-distilled_2026}. We therefore seek dense feedback on the agent's own decision contexts, focused where it currently struggles.

On-policy distillation (OPD) provides such dense feedback: a teacher scores the student's own generated responses, avoiding prefix mismatch~\citep{agarwal_-policy_2024}. On-policy self-distillation (OPSD) uses the same model as teacher, with additional privileged information~\citep{zhao_self-distilled_2026,wang_skill-sd_2026,lu_self-distilled_2026,yang2026agenticreinforcementlearningobservationcalibrated}. Selective intervention already helps allocate OPD across turns~\citep{zhou_sage-opd_2026}. We organize frontier-guided teaching around three design decisions:
\begin{itemize}
\item \textbf{Unfocused teaching locations.} A trajectory contains both reliable and difficult decisions. Applying supervision throughout it spends teaching effort on steps the agent already handles, while task-level returns offer little help in locating the bottleneck.
\item \textbf{Uncontrolled teaching scope.} A hindsight rule may help resolve one difficulty but lose relevance as the task advances. Extending its supervision across the remaining trajectory can interfere with later decisions; its strength and duration should follow its actual influence.
\item \textbf{Changing teaching needs.} Learning changes which steps the agent can execute reliably. Teaching positions and rules must follow the updated agent's difficulties, so that supervision shifts toward the next unmastered stage.
\end{itemize}

We also give the base model a successful trajectory from the same task as a reference and examine how its action predictions change (Figure~\ref{fig:intro-observations}(b)). The reference affects different trajectory positions unevenly, motivating control over both where teaching starts and how far it extends. Appendix~\ref{app:intro-observations} gives the diagnostic protocol.

An \emph{error frontier} is a position on the agent's own trajectory that separates a \emph{mastered prefix} from an \emph{unmastered suffix} (Section~\ref{sec:formulation}). Before this position, the current agent can execute the preceding steps consistently and with high success. From this position onward, it has yet to master the remaining steps: execution becomes less reliable, and successful progress is less likely. Earlier steps need little extra guidance, and excessive rules can interfere with correct decisions; we therefore start teaching at this boundary. Our insight is to use hindsight at this boundary to help the agent master the next stage. We propose \method{} to combine frontier localization with local self-distillation.

We first choose where to teach (\textbf{WHERE}). Action entropy provides an inexpensive way to recall uncertain decisions; we then compare support for sampled alternatives with support for the recorded action to select a teaching position. Entropy supplies candidates, while the support comparison helps distinguish a questionable choice from uncertainty among reasonable alternatives. Next, the same agent extracts a short guidance rule from recorded experience (\textbf{WHAT}), such as checking the requested color before choosing a product option. We use its rule-guided predictions as local self-distillation targets, turning hindsight into supervision on the agent's original responses (Figure~\ref{fig:intuition}).

We then decide how strongly and how long to teach (\textbf{HOW}). The rule's influence on predictions controls distillation weights and the extent of the teaching interval, concentrating supervision where the rule affects the agent's decisions. Full-trajectory GRPO continues alongside local distillation, so the agent also learns from task outcomes. Together, these components address where teaching begins and how far it extends.

As the agent masters more steps, its mastered prefix can extend into the previously unmastered suffix. We therefore repeat localization and teaching on fresh trajectories. We first initialize rule generation with externally labeled examples through SFT. Thereafter, the same agent acts, identifies teaching positions, writes rules, and learns from them; the updated agent supplies the next round's experience and rules. This cross-round loop is our \emph{recursive self-improvement} process (Section~\ref{sec:rsi}), which adapts teaching to changing bottlenecks. We evaluate the resulting agent without guidance.

We evaluate Qwen3-1.7B and Qwen3-4B on WebShop, ALFWorld, MiniHack, and ScienceWorld, with no guidance at test time. FrontierSD improves WebShop, ALFWorld seen, and MiniHack success at both scales. For Qwen3-4B, WebShop and ALFWorld seen success reach 78.0\% and 93.6\%, exceeding the strongest reported baselines by 4.3 and 10.8 percentage points, respectively. The learned behavior also transfers to unseen ALFWorld tasks, where 4B reaches 91.8\% success. Training curves show how the gains develop, while later teaching positions and fewer guidance triggers connect learning progress with an expanding mastered prefix. Ablations examine the localization and supervision choices behind these gains (Section~\ref{sec:experiments}).

Our contributions are threefold:
\begin{enumerate}
\item \textbf{A capability boundary for targeted teaching.} We define the error frontier as the position separating a mastered prefix from an unmastered suffix. This view explains why teaching should begin near the agent's current bottleneck: earlier steps already need little help, while the remaining steps offer room for learning.
\item \textbf{Frontier-guided local self-distillation.} We develop WHERE, WHAT, and HOW to recall and refine candidate positions, extract rules from hindsight, and control local distillation through the rule's influence. We combine this teaching with full-trajectory GRPO and repeat it on fresh experience, allowing the same agent to adapt its teaching as it learns.
\item \textbf{Evaluation of performance and teaching behavior.} We evaluate two model scales across four environments without guidance at test time. Main and unseen results assess the learned policy's task performance; ablations examine localization and supervision choices. Learning curves and frontier dynamics further show how the benefit develops and how teaching needs change during training.
\end{enumerate}

\section{Related Work}

\subsection{On-Policy Self-Distillation}

On-policy distillation supervises the learner's own generated responses~\citep{agarwal_-policy_2024,gu_minillm_2026,zhao2026opod}. OPSD obtains teacher targets from the same model under privileged context~\citep{zhao_self-distilled_2026}. That context may come from offline guidance~\citep{shi_skill1_2026,huang_skill-conditioned_2026,lu_self-distilled_2026} or trajectory-derived experience~\citep{wang_skill-sd_2026,yang_opid_2026,liu_procedural_2026,wu_seed_2026,liang_search-e1_2026}.

These signals enter optimization in different ways: as auxiliary distillation beside RL~\citep{lu_self-distilled_2026,tan_self-supervised_2026,wu_seed_2026,dong2026agentreflex}, or through redesigned outcome-based advantages~\citep{yang_self-distilled_2026,xie_evisd_2026,zhang2026stepopsdstepawareonlinepreference,wang2026agentopsdrecursiveselfdistillationagentic}. Confidence gates and evidence-dependent weighting address unreliable supervision~\citep{lu_self-distilled_2026,xia2026scoringactingoutcomeverifiedcomparative,liu2026privilegedguidancemisalignsstatematched}. SAGE-OPD selects turn-level intervention using environment feedback and teacher judgment, then weights supervision by teacher confidence~\citep{zhou_sage-opd_2026}. We instead couple a selected teaching origin with an influence-controlled interval. Our recursion refreshes experience and guidance across training rounds; AgentOPSD updates Bayesian beliefs across turns for credit assignment~\citep{wang2026agentopsdrecursiveselfdistillationagentic}.

Like FrontierSD, SEED refreshes a shared actor/analyzer after SFT and distills trajectory-level hindsight through confidence-gated shifts in sampled-token log-probabilities~\citep{wu_seed_2026}. We localize a teaching origin through entropy recall and sampled action support, extract a rule for that difficulty, and use JS influence to control both the weights and endpoint of local full-vocabulary forward-KL distillation.

\subsection{Process Supervision and Error Localization}

Process supervision connects outcomes to intermediate decisions. Let's Verify Step by Step learns from human step labels~\citep{lightman2023verify}; Math-Shepherd constructs step-wise supervision automatically~\citep{wang2024mathshepherd}. OmegaPRM combines binary search and Monte Carlo continuations to locate reasoning errors~\citep{DBLP:journals/corr/abs-2406-06592}. ProcessBench evaluates earliest-error identification against expert annotations~\citep{DBLP:conf/acl/ZhengZZLLYLZL25}.

Interactive tasks also require judging how actions advance a goal. AgentPRM models step-wise promise and progress~\citep{xi2025agentprmprocessrewardmodels}. Who\&When attributes multi-agent failures to an agent and decisive step~\citep{zhang2025whowhen}. Agent-R selects revision points in failed interactions and splices in MCTS-derived recovery paths for self-training~\citep{yuan2025agentrtraininglanguagemodel}. We use localization to allocate teaching: entropy recalls candidates, sampled action support selects a position, and a local rule supplies distillation targets on the recorded responses, without branching the environment for recovery search.

\section{Error Frontier}
\label{sec:formulation}

We define the \emph{error frontier} as a position on the agent's own trajectory separating a \emph{mastered prefix} from an \emph{unmastered suffix}. Before this position, the current agent can execute the steps consistently and with high success. From this position onward, it has yet to master the remaining steps, so continued progress is less reliable. For example, an agent may reliably search for products and open relevant pages, then struggle to select the correct option. The transition into that difficulty is its current frontier.

For task $x$, a guidance-free policy $\pi_\theta$ generates
\[
\tau=(s_1,y_1,a_1,o_1,\ldots,s_T,y_T,a_T,o_T).
\]
Here $T$ is the number of recorded decisions, $s_t$ the policy-visible state before decision $t$, $y_t$ the response, $a_t$ its parsed action, and $o_t$ the environment feedback. We retain decisions without an executable action. Let $h_t$ be the available history and $C_t=(h_t,s_t)$ the decision context. If the frontier is at $f$, then $\tau_{<f}$ is the mastered prefix and $\tau_{\geq f}$ the unmastered suffix. The boundary depends on the current policy and task. Section~\ref{sec:where} estimates a teaching position from observable action support, using it as a proxy for this capability boundary.

\paragraph{Why start teaching here?}
The mastered prefix already contains steps the agent can perform reliably, so it needs less additional guidance there. Excessive or irrelevant rules can perturb these correct decisions and introduce misleading supervision. We therefore concentrate extra teaching at the transition into the unmastered suffix, where the agent needs help with the next difficulty. Full-trajectory GRPO continues to train all decisions from task outcomes.

As the agent learns previously unreliable steps, the mastered prefix can extend and the frontier move toward later stages. We call this \emph{pushing the frontier}. Learning can also change the agent's path and the difficulties it encounters, so we locate the frontier again on fresh trajectories. The next section explains how we do this at manageable cost, what guidance we provide, and how we turn it into learning.

\section{Method}
\label{sec:method}

\subsection{Overview and Recursive Training}
\label{sec:iteration}
\label{sec:rsi}

\begin{figure}[!htbp]
    \centering
    \includegraphics[trim={10.169pt 28.689pt 11.328pt 9.002pt},clip,width=\textwidth]{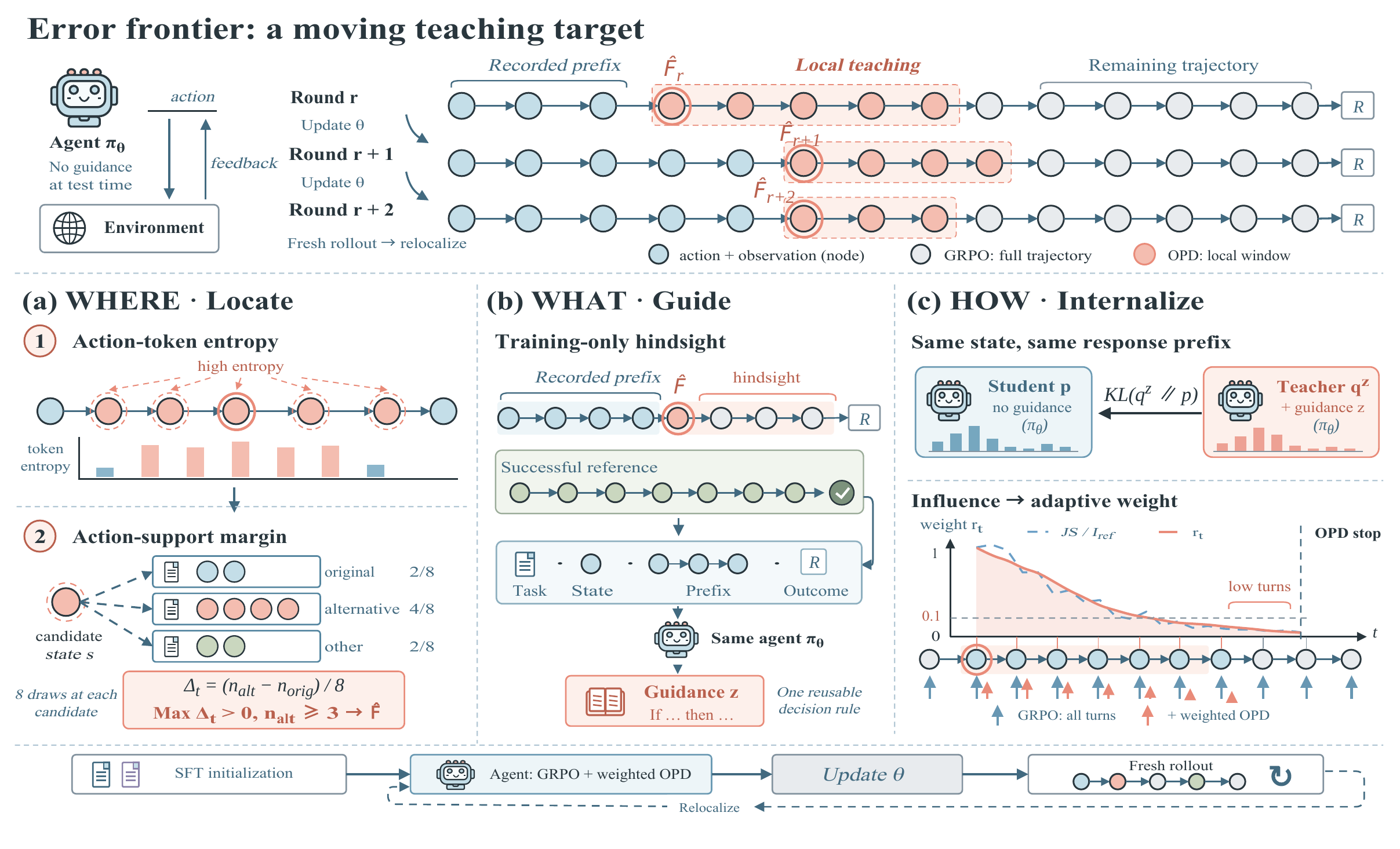}
    \caption{\method{} overview. WHERE selects a teaching position, WHAT extracts a hindsight rule with an optional successful reference, and HOW controls local OPD through JS influence. Complete-trajectory GRPO and local teaching jointly update the agent, which generates fresh experience and rules in the next round.}
    \label{fig:overview}
\end{figure}

We use the frontier as the starting point for a recurring teaching process (Figure~\ref{fig:overview}). Each round, we collect eight guidance-free rollouts per task, identify where the agent needs help (\textbf{WHERE}), extract a rule from what happened there (\textbf{WHAT}), and learn from that rule through local distillation (\textbf{HOW}). We retain full-trajectory GRPO so the agent also learns from task outcomes.

\paragraph{One agent for interaction and teaching.}
We want guidance to come from the agent's own experience as its capabilities change. We first use GLM-5.3 labels for SFT to initialize rule extraction. Thereafter, one trainable policy $\pi_\theta$ acts, samples localization alternatives, writes rules, and scores its recorded responses with and without a rule. The guided scoring branch shares the current weights and stops gradients; a frozen SFT copy supplies the KL reference.

\paragraph{Recursive self-improvement across rounds.}
At round $r$, we use $\theta_r$ to collect trajectories, choose teaching positions, and generate rules. Each rule is written once for its source trajectory and kept fixed during learning; teacher distributions are recomputed with current parameters in each minibatch. The joint update produces $\theta_{r+1}$, which supplies the next round's experience and rules. This feedback loop is what we mean by recursive self-improvement. We continue to use environment feedback and permitted training information, including ALFWorld training labels and PDDL success conditions. Evaluation removes guidance.

\subsection{WHERE · Locate}
\label{sec:where}

Following the frontier definition, we first need to locate where the agent's reliable behavior gives way to difficulty. We must repeat this search over many trajectories during training, so sampling alternatives at every decision would be costly. Entropy provides a cheap first signal, while additional samples let us check the agent's support for a particular action more closely. To balance localization quality and cost, we use a recall-and-rerank design: spend little computation to find candidates, then spend the sampling budget on these candidates.

\paragraph{Recall with action entropy.}
We first recall potentially uncertain positions from the recorded responses. Let $c_{t,k}=(x,h_t,s_t,y_{t,<k})$ denote the recorded input and response-token prefix. We compute
\begin{equation}
H_t=\frac{1}{|\mathcal D_t|}
 \sum_{k\in\mathcal D_t}
 \left[-\sum_v \pi_\theta(v\mid c_{t,k})
                  \log \pi_\theta(v\mid c_{t,k})\right],
\label{eq:entropy}
\end{equation}
where $\mathcal D_t$ contains the parsed action's token positions. The five highest-entropy decisions form $\mathcal C_5(\tau)$, with earlier decisions preferred on ties. When no action is parsed, recall uses the non-special response tokens.

\paragraph{Rerank with sampled action support.}
We then ask whether the agent consistently supports the action it actually took. At each recalled position, we draw eight independent next responses from the recorded input without executing their actions. We group the valid commands and match the original action to these groups. This converts the inexpensive token-level signal into action-level evidence: if another action attracts more support than the original, the recorded choice is less reliably supported by the agent itself. Let $n_t^{\rm orig}$ be its group's count, or zero if unmatched, and $n_t^{\rm alt}$ the largest count among other groups. Counts use only the eight new samples. We define
\begin{equation}
\begin{aligned}
\Delta_t&=(n_t^{\rm alt}-n_t^{\rm orig})/8,\\
\mathcal E(\tau)&=\{t\in\mathcal C_5(\tau):
             \Delta_t>0,\ n_t^{\rm alt}\geq3\},\\
\widehat F(\tau)&=\operatorname*{arg\,max}_{t\in\mathcal E(\tau)}
                      (\Delta_t,-t).
\end{aligned}
\label{eq:operational-frontier}
\end{equation}
We check all candidates and choose the largest positive margin with at least three votes for an alternative action, preferring earlier decisions on ties. If $\mathcal E(\tau)$ is empty, we set $\widehat F(\tau)=\bot$ and omit OPD for that trajectory. Otherwise, we begin teaching at position $\widehat F(\tau)$. This operational estimate prioritizes the strongest support discrepancy among recalled candidates. Sample support measures relative policy preference; task outcomes and ablations assess the usefulness of teaching there. Training analyses track selected positions as a proxy for the changing boundary.

We keep the denominator at eight even when some samples are invalid, so support counts remain comparable across candidates. Ambiguous original-to-group assignments abstain. Commands are normalized and grouped by exact identity, except WebShop search queries, which use deterministic task-visible lexical grouping (Appendix~\ref{app:localization}). Both successful and failed trajectories are eligible. Thus localization requires at most forty short model generations per trajectory, with no environmental branches.

\subsection{WHAT · Guide}
\label{sec:what}
\label{sec:guidance-sft}

Having selected where to teach, we next ask what could help the agent there. The final reward alone gives little instruction, whereas the recorded experience shows how its decisions developed. We therefore let the same agent review this experience and extract guidance for the selected difficulty. At $f=\widehat F(\tau)$, we build $X_f$ from the task, original decision input, complete compact interaction prefix, the recorded decisions from $f$ through $\min(f+3,T)$ (including $f$, at most four decisions), and episode outcome. We optionally add a successful rollout of the same task. The localization samples serve only to select the position; rule extraction uses the recorded trajectory evidence.

We compress this experience into a reusable condition--decision rule, sampling $z\sim\pi_\theta(\cdot\mid X_f)$ from the same policy. Such a rule says when the guidance applies and what to check next, making the hindsight useful for learning at the frontier. We preserve each environment's information boundary and distinguish current observations from hindsight. Well-formed rules enter guided scoring; unusable generations leave the trajectory's GRPO supervision active.

To make the agent capable of producing this guidance, we first train rule extraction through SFT with GLM-5.3 labels. For Qwen3-4B ALFWorld and WebShop, offline examples reuse the online localization and hindsight interface to teach the operation needed during RL (Appendix~\ref{app:sft}). We apply cross-entropy only to rule-completion tokens, then use the resulting policy for acting and rule generation. Online rules are sampled from the current agent without a separate rule-generation loss (Appendix~\ref{app:sft}).

\subsection{HOW · Internalize}
\label{sec:how}
\label{sec:prefix-opd}

We now know where to teach and what guidance to provide. The remaining question is how to turn this guidance into a policy that can act without the rule. We use the same agent's rule-guided predictions as teacher targets for its original responses. At each token, the ordinary input gives the student distribution, and adding $z$ gives a stop-gradient teacher:
\begin{equation}
\begin{aligned}
p_{t,k}&=\pi_\theta(\cdot\mid c_{t,k}),\\
q^z_{t,k}&=\pi_{\sg(\theta)}(\cdot\mid c_{t,k},z).
\end{aligned}
\label{eq:prefix-alignment}
\end{equation}
Both branches use the same recorded history, state, response-token prefix $y_{t,<k}$, and current parameters; the guided branch adds $z$ and stops gradients. Distillation transfers the rule-induced change into the unguided policy.

\paragraph{Influence and teaching strength.}
We also need to decide how strongly and how long to apply this supervision. A rule extracted for the current difficulty may matter for nearby decisions and lose relevance as the trajectory moves on. We therefore use its effect on predictions to control teaching, measuring influence with full-vocabulary Jensen--Shannon divergence:
\begin{equation}
I_t=\frac{1}{|\mathcal J_t|}
 \sum_{k\in\mathcal J_t}\mathrm{JS}
       \big(q^z_{t,k},\sg(p_{t,k})\big).
\label{eq:influence}
\end{equation}
where $\mathcal J_t$ contains action-token positions, falling back to all saved response tokens, including special tokens, when no action is parsed. Influence controls teaching strength and duration; the distillation loss supervises the full response.

Teaching starts only when the recorded history and visible state match $C_f$. We initialize $\bar I_f=I_f$ and smooth later influences with $\bar I_t=0.5I_t+0.5\bar I_{t-1}$. Up to the first three guided decisions establish a reference scale:
\[
b_t=\max_{f\leq j\leq\min(t,f+2)}\bar I_j,\qquad
r_t=\min\left(1,\frac{\bar I_t}{b_t+\epsilon}\right).
\]
We let this reference grow during the first three guided decisions, then fix it at $I_{\rm ref}$. We use $\epsilon>0$ for numerical stability. Scaling influence relative to this reference lets us adapt teaching strength to each rule without choosing a shared absolute JS threshold.

\paragraph{Teaching interval and local objective.}
We want the teaching interval to end when the rule's influence stays small, while tolerating an isolated low-influence decision. We therefore stop after two consecutive decisions with $r_t<0.1$. Let $e$ be the first such stopping decision, or the trajectory's last decision if none occurs. The active interval is $[f,e]$; the stopping decision receives its computed weight, after which the interval remains closed.

Within this interval, we train the unguided policy to match the guided teacher, minimizing forward KL over all original response tokens:
\begin{equation}
\begin{aligned}
\ell_{i,t}^{\rm OPD}
 &=\frac{1}{|y_{i,t}|}\sum_k
       \mathrm{KL}(q^z_{i,t,k}\Vert p_{i,t,k}),\\
\mathcal L_{\rm OPD}
 &=\frac{1}{N_{\rm orig}}\sum_{i,t}
       \sg(w_{i,t})\,\ell_{i,t}^{\rm OPD}.
\end{aligned}
\label{eq:objective}
\end{equation}
Here $i$ indexes trajectories, $N_{\rm orig}=\sum_iT_i$, and $w_{i,t}=r_{i,t}$ inside the interval and zero outside. We normalize by all original decisions, counted before applying the teaching mask. Shorter intervals and smaller weights therefore reduce the total teaching contribution. We stop gradients through the teacher targets, influence, and weights, and update only the unguided student distribution. We assess the resulting policy without guidance.

\begin{table}[t]
\centering
\caption{Main results (mean $\pm$ standard deviation). ALFWorld/WebShop baseline results are taken from Table~1 of OCSD~\citep{yang2026agenticreinforcementlearningobservationcalibrated}. Success rates are percentages; WebShop score uses 0--100. Darker/lighter blue marks the best/second-best mean within each backbone, including ties.}
\label{tab:ocsd_qwen3_1.7b_4b}
\setlength{\tabcolsep}{3pt}
\renewcommand{\arraystretch}{1.12}
\resizebox{\textwidth}{!}{%
\begin{tabular}{lccccccc|cc|c|cc}
\toprule
\multirow{2}{*}{Method} & \multicolumn{7}{c|}{ALFWorld (Seen)} & \multicolumn{2}{c|}{WebShop} & MiniHack & \multicolumn{2}{c}{ScienceWorld} \\
\cmidrule(lr){2-8}\cmidrule(lr){9-10}\cmidrule(lr){11-11}\cmidrule(lr){12-13}
& Pick & Look & Clean & Heat & Cool & Pick2 & Overall & Score & Success & Success & Score & Success \\
\midrule
\multicolumn{13}{c}{\textbf{Qwen3-1.7B}} \\
\midrule
Vanilla & $19.2{\pm}1.1$ & $32.7{\pm}4.0$ & $5.1{\pm}1.5$ & $1.5{\pm}0.9$ & $3.9{\pm}1.1$ & $5.8{\pm}1.3$ & $10.3{\pm}0.8$ & $36.6{\pm}1.3$ & $4.7{\pm}1.9$ & $23.4{\pm}1.4$ & $-31.7{\pm}1.2$ & $1.3{\pm}0.5$ \\
Skill-prompt & $15.2{\pm}1.8$ & $28.8{\pm}4.3$ & $4.5{\pm}0.6$ & $0.6{\pm}0.6$ & $3.3{\pm}1.0$ & $8.8{\pm}1.2$ & $9.3{\pm}0.7$ & $37.9{\pm}0.9$ & $6.3{\pm}2.1$ & $21.9{\pm}3.1$ & $-28.0{\pm}3.2$ & $1.6{\pm}0.0$ \\
GRPO & $70.4{\pm}8.0$ & $44.0{\pm}1.7$ & $40.6{\pm}7.4$ & $38.6{\pm}5.8$ & $33.5{\pm}5.8$ & $30.1{\pm}7.3$ & $46.6{\pm}0.7$ & $71.8{\pm}1.0$ & $48.4{\pm}0.8$ & $52.3{\pm}4.3$ & $8.3{\pm}3.1$ & $23.4{\pm}1.4$ \\
OPSD & $33.9{\pm}3.5$ & $30.7{\pm}7.9$ & $5.4{\pm}1.8$ & $1.6{\pm}2.3$ & $2.9{\pm}2.0$ & $11.6{\pm}5.0$ & $14.6{\pm}0.3$ & $50.3{\pm}0.6$ & $11.4{\pm}0.5$ & $38.8{\pm}2.3$ & $6.3{\pm}2.9$ & $17.2{\pm}2.7$ \\
GRPO+OPSD & $63.1{\pm}7.2$ & $41.5{\pm}5.8$ & $39.2{\pm}9.2$ & $44.8{\pm}9.4$ & $27.3{\pm}7.5$ & $29.1{\pm}1.8$ & $41.7{\pm}2.4$ & $72.1{\pm}2.6$ & $49.0{\pm}1.9$ & $43.0{\pm}4.3$ & $9.7{\pm}2.3$ & $24.5{\pm}1.6$ \\
RLSD & $49.9{\pm}6.2$ & $51.9{\pm}2.6$ & $46.7{\pm}3.6$ & $47.5{\pm}7.4$ & $38.1{\pm}3.0$ & $26.9{\pm}4.7$ & $44.5{\pm}1.7$ & \cellcolor{MainSecond}$76.3{\pm}2.5$ & $52.1{\pm}2.9$ & $63.0{\pm}3.7$ & \cellcolor{MainBest}$\mathbf{13.2{\pm}3.1}$ & $25.0{\pm}2.7$ \\
SDAR & $66.7{\pm}9.7$ & $50.6{\pm}7.5$ & $51.4{\pm}7.9$ & $33.7{\pm}6.5$ & $43.1{\pm}3.0$ & $38.1{\pm}3.6$ & $51.6{\pm}2.2$ & $73.4{\pm}1.0$ & $53.4{\pm}1.0$ & \cellcolor{MainSecond}$65.1{\pm}2.7$ & $8.9{\pm}1.3$ & \cellcolor{MainSecond}$27.3{\pm}1.4$ \\
OCSD & \cellcolor{MainSecond}$75.9{\pm}4.9$ & \cellcolor{MainSecond}$63.1{\pm}2.7$ & \cellcolor{MainSecond}$51.9{\pm}4.7$ & \cellcolor{MainSecond}$47.7{\pm}4.3$ & \cellcolor{MainSecond}$46.4{\pm}1.8$ & \cellcolor{MainSecond}$42.3{\pm}6.9$ & \cellcolor{MainSecond}$55.5{\pm}2.3$ & $74.7{\pm}3.4$ & \cellcolor{MainSecond}$54.4{\pm}1.2$ & $64.3{\pm}3.2$ & \cellcolor{MainSecond}$10.9{\pm}0.4$ & \cellcolor{MainBest}$\mathbf{29.4{\pm}1.2}$ \\
\textbf{\method{} (Ours)} & \cellcolor{MainBest}$\mathbf{80.0{\pm}2.9}$ & \cellcolor{MainBest}$\mathbf{69.2{\pm}0.0}$ & \cellcolor{MainBest}$\mathbf{85.2{\pm}3.7}$ & \cellcolor{MainBest}$\mathbf{81.3{\pm}0.0}$ & \cellcolor{MainBest}$\mathbf{64.0{\pm}4.0}$ & \cellcolor{MainBest}$\mathbf{50.0{\pm}4.2}$ & \cellcolor{MainBest}$\mathbf{72.1{\pm}1.2}$ & \cellcolor{MainBest}$\mathbf{77.7{\pm}1.8}$ & \cellcolor{MainBest}$\mathbf{65.8{\pm}1.1}$ & \cellcolor{MainBest}$\mathbf{75.8{\pm}3.4}$ & $10.4{\pm}0.5$ & \cellcolor{MainBest}$\mathbf{29.4{\pm}1.2}$ \\
\midrule
\multicolumn{13}{c}{\textbf{Qwen3-4B}} \\
\midrule
Vanilla & $62.6{\pm}1.1$ & $30.7{\pm}3.1$ & $20.7{\pm}1.3$ & $27.1{\pm}3.2$ & $9.5{\pm}3.3$ & $16.8{\pm}2.3$ & $29.6{\pm}3.1$ & $15.6{\pm}2.1$ & $6.3{\pm}1.1$ & $28.1{\pm}3.1$ & $-22.5{\pm}3.3$ & $8.6{\pm}2.7$ \\
Skill-prompt & $65.7{\pm}1.2$ & $14.4{\pm}2.0$ & $32.9{\pm}0.8$ & $28.1{\pm}4.3$ & $13.5{\pm}4.1$ & $15.1{\pm}5.1$ & $32.3{\pm}0.8$ & $11.8{\pm}4.5$ & $7.6{\pm}0.4$ & $40.9{\pm}4.6$ & $-13.0{\pm}2.1$ & $11.7{\pm}3.4$ \\
GRPO & $86.6{\pm}0.6$ & $59.5{\pm}5.1$ & $77.7{\pm}4.7$ & $67.1{\pm}4.1$ & $55.5{\pm}4.1$ & $55.0{\pm}6.2$ & $70.6{\pm}1.3$ & $81.9{\pm}3.6$ & $69.5{\pm}1.5$ & $77.1{\pm}3.3$ & $3.2{\pm}2.7$ & $41.4{\pm}1.6$ \\
OPSD & $69.9{\pm}3.1$ & $39.2{\pm}4.5$ & $26.6{\pm}4.8$ & $26.6{\pm}4.3$ & $19.0{\pm}4.6$ & $26.4{\pm}3.8$ & $36.2{\pm}1.9$ & $27.8{\pm}2.9$ & $4.7{\pm}0.6$ & $43.5{\pm}3.2$ & $10.3{\pm}2.9$ & $23.2{\pm}3.2$ \\
GRPO+OPSD & $88.0{\pm}0.4$ & $52.8{\pm}3.9$ & $75.0{\pm}4.1$ & $62.5{\pm}3.4$ & $68.3{\pm}8.1$ & $36.9{\pm}2.5$ & $68.8{\pm}2.6$ & $82.5{\pm}1.0$ & $71.3{\pm}1.2$ & $55.2{\pm}4.0$ & $13.2{\pm}3.3$ & $41.9{\pm}2.7$ \\
RLSD & $87.9{\pm}4.0$ & \cellcolor{MainSecond}$83.4{\pm}1.8$ & $80.5{\pm}0.9$ & $70.3{\pm}9.4$ & $79.6{\pm}4.0$ & $60.6{\pm}4.3$ & $79.2{\pm}1.0$ & $83.7{\pm}1.5$ & $68.5{\pm}1.2$ & $77.3{\pm}3.6$ & $20.1{\pm}1.9$ & \cellcolor{MainSecond}$45.3{\pm}2.8$ \\
SDAR & $85.2{\pm}4.8$ & $63.9{\pm}4.0$ & $76.8{\pm}5.8$ & $80.0{\pm}3.6$ & $72.9{\pm}4.4$ & $42.6{\pm}6.9$ & $73.2{\pm}1.3$ & $83.6{\pm}1.8$ & $71.9{\pm}0.7$ & $76.6{\pm}2.8$ & \cellcolor{MainSecond}$23.0{\pm}3.2$ & \cellcolor{MainBest}$\mathbf{46.4{\pm}3.2}$ \\
OCSD & \cellcolor{MainSecond}$91.9{\pm}1.4$ & $77.2{\pm}4.4$ & \cellcolor{MainSecond}$90.2{\pm}4.6$ & \cellcolor{MainSecond}$84.7{\pm}4.4$ & \cellcolor{MainSecond}$80.9{\pm}4.3$ & \cellcolor{MainSecond}$63.8{\pm}0.9$ & \cellcolor{MainSecond}$82.8{\pm}0.7$ & \cellcolor{MainSecond}$86.9{\pm}1.7$ & \cellcolor{MainSecond}$73.7{\pm}1.6$ & \cellcolor{MainSecond}$81.0{\pm}1.8$ & $16.36{\pm}1.8$ & $36.7{\pm}4.3$ \\
\textbf{\method{} (Ours)} & \cellcolor{MainBest}$\mathbf{100.0{\pm}0.0}$ & \cellcolor{MainBest}$\mathbf{84.6{\pm}0.0}$ & \cellcolor{MainBest}$\mathbf{96.3{\pm}3.7}$ & \cellcolor{MainBest}$\mathbf{100.0{\pm}0.0}$ & \cellcolor{MainBest}$\mathbf{92.0{\pm}4.0}$ & \cellcolor{MainBest}$\mathbf{83.3{\pm}4.2}$ & \cellcolor{MainBest}$\mathbf{93.6{\pm}1.2}$ & \cellcolor{MainBest}$\mathbf{87.4{\pm}1.8}$ & \cellcolor{MainBest}$\mathbf{78.0{\pm}1.2}$ & \cellcolor{MainBest}$\mathbf{88.0{\pm}4.0}$ & \cellcolor{MainBest}$\mathbf{33.4{\pm}3.3}$ & \cellcolor{MainSecond}$45.3{\pm}2.1$ \\
\bottomrule
\end{tabular}
}
\end{table}

\subsection{Joint Objective with Complete-Trajectory GRPO}
\label{sec:grpo}

Local guidance helps with the current difficulty, while task outcomes tell us whether the whole interaction succeeded. We therefore combine both signals. We assign each recorded decision its episode return with the invalid-action penalty, then normalize across the decision rows of the task's eight rollouts. Each decision's response tokens share this advantage. We use clipped policy optimization with negative-advantage dual clipping, a frozen-reference KL penalty, and entropy regularization:
\begin{equation}
\begin{aligned}
\mathcal L_{\rm agent}
 &=\mathcal L_{\rm PG}+\beta_{\rm KL}\mathcal L_{\rm refKL}\\
 &\quad-\beta_{\rm H}\mathcal H
       +\lambda_{\rm OPD}\mathcal L_{\rm OPD}.
\end{aligned}
\label{eq:agent-loss}
\end{equation}
We apply the policy-gradient, reference, and entropy terms to all original decisions, including those outside teaching intervals or without a usable rule. We restrict only $\mathcal L_{\rm OPD}$ to the adaptive interval. We freeze the RL-initial SFT checkpoint as the reference to anchor regularization, while using the current agent as the self-teacher so that guidance evolves with learning.

We apply the joint update and synchronize the resulting parameters for the next round of interaction and rule generation. Localization and guidance-conditioned scoring reuse recorded experience; additional online environment interaction comes from the next guidance-free rollout batch. Coefficients, minibatch normalization, optimizer settings, and generation budgets are given in Appendix~\ref{app:hparams}.

\section{Experiments}
\label{sec:experiments}

We evaluate FrontierSD without guidance to assess task performance, transfer, and progress during training. We then examine teaching location, influence weighting within the adaptive interval, and changes in selected positions during learning.

\subsection{Experimental Setup}
\label{sec:experimental-setup}

\textbf{Models and benchmarks.} We evaluate Qwen3-1.7B and Qwen3-4B~\citep{qwen2025qwen3} on WebShop~\citep{yao2023webshopscalablerealworldweb} for product search and purchase, ALFWorld~\citep{shridhar2021alfworld} for household instruction following, MiniHack~\citep{samvelyan2021minihack} for navigation and object interaction, and ScienceWorld~\citep{wang2022scienceworld} for interactive scientific reasoning. We report ALFWorld seen/unseen separately. Benchmark descriptions and metrics appear in Appendix~\ref{app:benchmarks}.

\textbf{Baselines.} We compare the original Qwen3 backbone (Vanilla)~\citep{qwen2025qwen3} and an inference-time guidance control (Skill-prompt), reward-only GRPO~\citep{shao2024deepseekmath}, direct self-distillation (OPSD~\citep{zhao_self-distilled_2026} and GRPO+OPSD~\citep{lu_self-distilled_2026}), and methods that integrate distillation with RL: RLSD~\citep{yang_self-distilled_2026}, SDAR~\citep{lu_self-distilled_2026}, and OCSD~\citep{yang2026agenticreinforcementlearningobservationcalibrated}. These comparisons cover prompting, outcome learning, and dense supervision; Appendix~\ref{app:baselines} describes each baseline.

\textbf{Training.} We initialize guidance generation by SFT with GLM-5.3 labels, then run 150 RL updates with 16 tasks and eight rollouts per task. Two-stage localization and adaptive OPD focus teaching on current bottlenecks (Appendix~\ref{app:hparams}).

\textbf{Evaluation protocol.} FrontierSD acts without guidance at test time; Skill-prompt retains inference-time guidance. We report repeated evaluations as mean and standard deviation across three seeds at a fixed checkpoint; ALFWorld seen uses sample standard deviation and pools all 140 tasks for each run's Overall. Baseline statistics follow their reported protocols. Our 4B ALFWorld and WebShop results use update 150; complete-split evaluations use temperature 0.4 and seed 0.

\subsection{Main Results}
\label{sec:main-results}

We compare task performance in Table~\ref{tab:ocsd_qwen3_1.7b_4b}, evaluating FrontierSD without guidance, and find three patterns.

\textbf{(1) Local teaching improves the agent's independent task performance.} On WebShop and ALFWorld, we outperform both outcome-only GRPO and trajectory-wide distillation baselines at both model scales (Table~\ref{tab:ocsd_qwen3_1.7b_4b}). For example, our 4B success exceeds the strongest reported baseline, OCSD, by 4.3 and 10.8 percentage points, respectively. The consistent direction across scales supports combining task-outcome learning with targeted teaching. Since evaluation removes guidance, the gains reflect improvements retained by the learned policy when acting independently.

\textbf{(2) The gains extend across different task requirements.} We obtain the highest reported mean in all six ALFWorld categories at both scales. At 1.7B, the largest gains over OCSD occur on Clean and Heat. Thus, the aggregate result reflects broad improvements rather than a single category dominating the mean. Trajectory-specific localization lets different tasks receive guidance near their respective difficulties.

\textbf{(3) Local teaching extends to diverse interaction tasks.} On MiniHack, FrontierSD improves success over the strongest reported baseline at both scales, extending its benefit beyond shopping and household tasks. On ScienceWorld, 1.7B ties the highest reported success rate, while 4B achieves the highest score. Both ScienceWorld metrics improve from 1.7B to 4B. This scale-dependent pattern is consistent with the environment's demanding scientific reasoning and interaction requirements, which may benefit from greater model capacity.

\subsection{Ablation Study}
\label{sec:ablations}
\label{sec:localization-analysis}
\label{sec:scope}

We compare Qwen3-1.7B variants to examine local teaching, its starting position, and its strength (Table~\ref{tab:ablation_1.7b}). ALFWorld Overall weights seen/unseen success by 140/134 tasks; changes use the displayed values.

\textbf{Local distillation adds useful supervision.} Removing OPD while retaining full-trajectory GRPO lowers performance on both environments, including a 9.5-point drop in ALFWorld Overall. This supports teaching intermediate decisions alongside outcome learning. The main-table GRPO uses a different configuration; this ablation isolates OPD within Table~\ref{tab:ablation_1.7b}.

\begin{table}[t]
\centering
\caption{Qwen3-1.7B ablations. ALFWorld Overall combines seen/unseen success; WebShop reports score/success. Subscripts show changes from the full method (success: percentage points).}
\label{tab:ablation_1.7b}
\small
\setlength{\tabcolsep}{2pt}
\begin{adjustbox}{max width=\linewidth}
\begin{tabular}{@{}lccccc@{}}
\toprule
\multirow{2}{*}{Variant} & \multicolumn{3}{c}{ALFWorld} & \multicolumn{2}{c}{WebShop} \\
\cmidrule(lr){2-4}\cmidrule(lr){5-6}
& Seen & Unseen & Overall & Score & Success \\
\midrule
Random localization & $62.9\abldown{9.2}$ & $64.2\abldown{1.5}$ & $63.5\abldown{5.5}$ & $66.7\abldown{11.0}$ & $52.2\abldown{13.6}$ \\
First-step localization & $67.8\abldown{4.3}$ & $65.7\ablsame$ & $66.8\abldown{2.2}$ & $68.1\abldown{9.6}$ & $53.8\abldown{12.0}$ \\
LLM-judged localization & $71.2\abldown{0.9}$ & $66.0\ablup{0.3}$ & $68.7\abldown{0.3}$ & $75.2\abldown{2.5}$ & $64.7\abldown{1.1}$ \\
Max-entropy localization & $69.3\abldown{2.8}$ & $64.7\abldown{1.0}$ & $67.1\abldown{1.9}$ & $75.3\abldown{2.4}$ & $63.2\abldown{2.6}$ \\
w/o OPSD & $61.4\abldown{10.7}$ & $57.5\abldown{8.2}$ & $59.5\abldown{9.5}$ & $70.8\abldown{6.9}$ & $62.3\abldown{3.5}$ \\
w/o influence weighting & $68.3\abldown{3.8}$ & $63.2\abldown{2.5}$ & $65.8\abldown{3.2}$ & $77.2\abldown{0.5}$ & $65.1\abldown{0.7}$ \\
\midrule
\rowcolor{AblFull}
\textbf{\method{} (full)} & \textbf{72.1} & \textbf{65.7} & \textbf{69.0} & \textbf{77.7} & \textbf{65.8} \\
\bottomrule
\end{tabular}
\end{adjustbox}
\end{table}

\textbf{Where teaching starts matters.} We replace frontier selection with the first decision or a uniformly sampled position. First-step extraction reads the full suffix; random extraction uses four-step hindsight, so these controls vary both placement and extraction. Both reduce ALFWorld Overall and WebShop performance; first-step localization ties the full method on unseen tasks. WebShop success drops by up to 13.6 points, favoring teaching near the current bottleneck over a fixed or random starting point.

\textbf{Action support refines entropy candidates.} We compare two-stage localization with direct Qwen judgment and max-entropy selection. Our selector recalls five high-entropy positions and samples eight next responses per position to compare alternative and recorded-action support. It improves ALFWorld Overall and both WebShop metrics over either control, supporting sampled refinement after entropy recall. The full selector and direct judgment give similar unseen success, while sampled refinement improves the aggregate comparison.

\textbf{Teaching strength should follow rule influence.} Keeping adaptive stopping, we set all active distillation weights to one. ALFWorld performance falls more clearly than WebShop performance, supporting influence-based weighting within the selected interval. The uneven influence in Figure~\ref{fig:intro-observations}(b) offers a consistent motivation.

The OPD-weight sweep favors a small coefficient, supporting a balance between local teaching and outcome learning (Appendix~\ref{sec:opd-weight-sensitivity}, Table~\ref{tab:opd-weight-sensitivity}).

\subsection{Generalization and Learning Efficiency}
\label{sec:unseen}

\begin{figure}[!htbp]
\begin{minipage}[t]{0.485\textwidth}\vspace{0pt}
\centering
\captionsetup{font=papersubcaption}
\includegraphics[width=\linewidth]{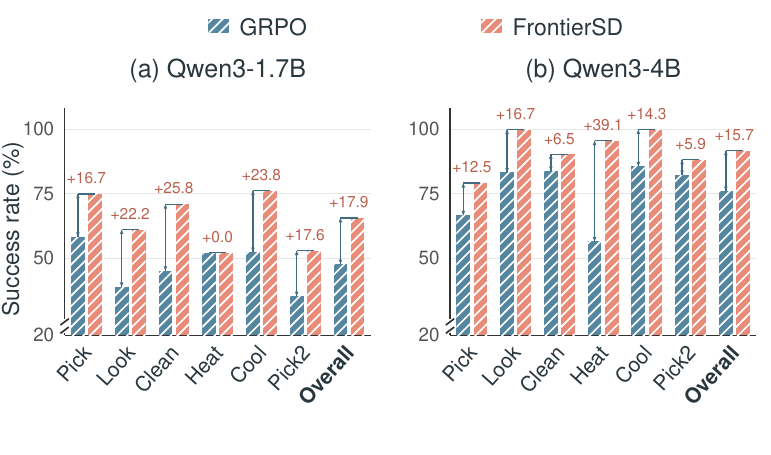}
\caption{Complete-split ALFWorld unseen success: (a) Qwen3-1.7B; (b) Qwen3-4B. Coral/blue-gray denotes \method{}/GRPO; annotations give percentage-point differences. Both axes start at 20\%. Overall is episode-weighted.}
\label{fig:alfworld_unseen}
\end{minipage}\hfill
\begin{minipage}[t]{0.485\textwidth}\vspace{0pt}
\centering
\captionsetup{font=papersubcaption}
\includegraphics[width=\linewidth]{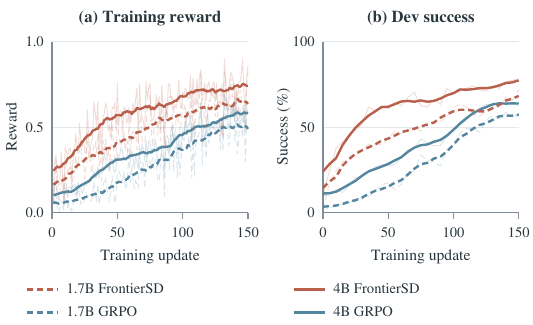}
\caption{WebShop reward (left) and dev success (right). Coral/blue-gray denotes \method{}/GRPO; dashed/solid lines denote 1.7B/4B. Light traces show observations; thick traces show centered $\pm10$-update means.}
\label{fig:training_dynamics}
\end{minipage}
\end{figure}

\textbf{The learned behavior transfers to unseen tasks.} We evaluate on all 134 ALFWorld unseen tasks and compare six task categories with GRPO (Figure~\ref{fig:alfworld_unseen}). The 4B result uses the same final-update-150 model and guidance-free protocol as the 140-task seen evaluation. Its unseen success remains close to seen performance and reaches 91.8\%, with gains over GRPO in every category. Thus, the benefit extends to new task instances across different interaction requirements, supporting the transfer of behavior learned through local teaching. At 1.7B, FrontierSD succeeds on 88 of 134 tasks (65.7\%), improving over GRPO in five categories and matching it on Heat. Table~\ref{tab:ablation_1.7b} uses this same 1.7B unseen result.

\textbf{The advantage emerges early and persists during training.} We track WebShop reward at each update and dev success every five updates from 0 to 150, for both methods and model scales (Figure~\ref{fig:training_dynamics}). Light traces retain individual observations; thick traces average observations within $\pm10$ updates. FrontierSD exceeds GRPO at every measured post-initialization dev evaluation at both scales. The 4B model surpasses GRPO's best dev result over its full run by update 65, showing that the advantage develops well before training ends. Reward and dev success improve together despite fluctuations, suggesting that the training benefit carries over to dev tasks. These curves compare progress per base-rollout update; FrontierSD additionally performs localization generations and paired scoring.

\subsection{Progression: How Do Teaching Positions Change during Training?}
\label{sec:frontier-dynamics}

\begin{wrapfigure}{r}{0.51\textwidth}
\vspace{-6pt}
\centering
\captionsetup{font=papersubcaption}
\includegraphics[width=\linewidth]{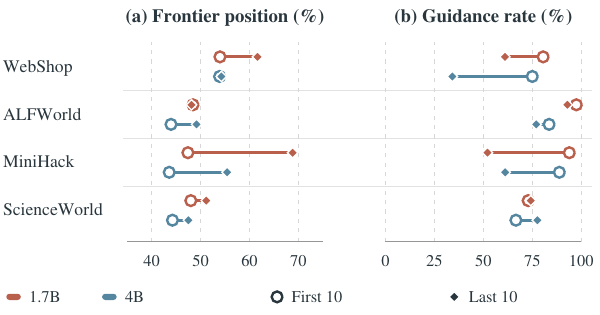}
\caption{Relative frontier position (left) and guidance rate (right; fraction with a selected position). Circles/diamonds show first-/last-ten-recorded-round means; coral/blue-gray denotes 1.7B/4B. Horizontal segments connect paired windows. Axis ranges are 35--75\% and 0--100\%.}
\label{fig:frontier-guidance-dynamics}
\end{wrapfigure}

We track all eight environment--model settings using two statistics (Figure~\ref{fig:frontier-guidance-dynamics}). Relative frontier position divides the selected turn by the actual length of its failed trajectory; a higher value places teaching farther into the trajectory. Guidance rate is the fraction of trajectories processed by WHERE for which two-stage localization selects a teaching position, $\widehat F(\tau)\ne\bot$. We compare the first ten and last ten available recorded rounds for each statistic, using each run's available observation window.

\textbf{Selected teaching positions move later.} Selected positions move later in seven of eight settings, with the largest shifts on MiniHack; ALFWorld 1.7B is nearly unchanged (Figure~\ref{fig:frontier-guidance-dynamics}(a)). This pattern is consistent with a growing mastered prefix: as earlier steps become more reliable, teaching shifts toward difficulties farther along the task. Relocalization lets us follow those difficulties instead of continuing to teach at an earlier position. The statistic describes changing failed trajectories and task mixtures, so we interpret the shift together with task performance and guidance demand.

\textbf{Later teaching positions usually coincide with fewer localization triggers.} Guidance rates decline in all WebShop, ALFWorld, and MiniHack settings (Figure~\ref{fig:frontier-guidance-dynamics}(b)); WebShop 4B falls from about 75\% to 34\%. Thus, fewer trajectories meet the action-support criterion for selecting a teaching position. Alongside later selected positions and improved guidance-free performance, this pattern is consistent with the agent handling more of the task independently. On ScienceWorld, guidance rates rise while selected positions move later, indicating that localization continues to identify teaching opportunities farther along the trajectory.

\begin{samepage}
\section{Conclusion}

We teach at the agent's current bottleneck and revisit it after learning. The error frontier separates a mastered prefix from an unmastered suffix, directing local self-distillation toward the next stage. After SFT, the same agent acts, localizes, writes rules, and supplies guided targets. Local teaching and full-trajectory GRPO update the agent for the next round, without extra online environment branches or teacher trajectories. We improve guidance-free WebShop and ALFWorld performance; across four environments, seven of eight settings show later frontiers and six show lower guidance rates.

\end{samepage}

\bibliographystyle{plainnat}
\bibliography{bibliography/related_work}

\clearpage
\appendix
\section{Implementation and Training Details}
\label{app:hparams}

\paragraph{Training task pools.}
We draw tasks from each environment's native training split. WebShop, ALFWorld, MiniHack, and ScienceWorld contain 2,120, 3,553, 1,000, and 3,194 training tasks, respectively. MiniHack tasks are defined by environment seeds.

\subsection{Offline Supervised Initialization}
\label{app:sft}

We prepare the agent for hindsight-guidance extraction using offline supervision. With the original same-scale Qwen model as source agent, we collect eight trajectories on each of 180 training tasks per benchmark, separating 162 tasks for SFT training and 18 for validation. We use both successful and failed experience. Offline position selection controls which examples teach this extraction interface; online WHERE separately determines which policy decisions receive OPD.

For supervised initialization on Qwen3-4B ALFWorld and WebShop, we reuse the online localization function: Top-5 recall, eight samples per candidate, and support-margin selection after all candidates are checked. 

We use GLM-5.3 for offline structured guidance labels and train only on guidance completion tokens:
\begin{equation}
\mathcal L_{\rm SFT}
=-\mathbb E_{(X,z^*)\sim\mathcal D_{\rm SFT}}
 \frac{1}{|z^*|}\sum_k
 \log\pi_\theta(z_k^*\mid X,z_{<k}^*).
\label{eq:guidance-sft}
\end{equation}
We perform three full-parameter epochs of guidance-completion training at learning rate $5\times10^{-6}$ and global batch size 8. The resulting weights initialize the agent and its frozen reference. Since this SFT changes all parameters, we distinguish its initialization effect from subsequent local distillation. Online self-teaching uses the resulting agent parameters.

\subsection{Action Grouping, Masks, and Guidance Budgets}
\label{app:localization}

For ALFWorld, MiniHack, and ScienceWorld, we group valid commands by exact match after case and whitespace normalization, preserving object identifiers and targets. We retain malformed or no-action responses in the records but exclude them from valid action groups; infrastructure failures remain sampling errors. For WebShop, we group non-search commands exactly and merge approximate \texttt{search[...]} intents using deterministic lexical rules grounded in visible requests and product-category vocabulary from training tasks. We use complete-link merging at threshold 0.5 with color, size, and numeric-conflict guards, without hidden target identifiers or executed search results.

We form groups from the eight new samples first, then attach the original action to these fixed groups. Only the eight new samples determine group membership and support counts. Unmatched originals have zero support.

A recorded action that matches multiple WebShop search groups is ineligible. We always use eight as the support-margin denominator; including when some samples are malformed. Localization seeds are bound to request identities. Online selection uses the support-margin eligibility rule in Section~\ref{sec:where}.

Entropy recall uses action tokens, falling back to non-special response tokens when no action is parsed. Influence uses the full-vocabulary JS on action tokens, falling back to all saved response tokens, including special tokens. Forward KL uses all original response tokens. These masks give each quantity its stated token-level aggregation.

For online guidance extraction, we retain the full observation prefix and up to four recorded decisions, from $f$ through $\min(f+3,T)$, including the selected decision. Rule generation uses one attempt, temperature 0.4, top-$p=1$, top-$k=0$, disabled Qwen thinking, and at most 512 output tokens. Extraction contexts are 32,768 for ALFWorld and the native environments and 12,288 for WebShop. We do not retry or truncate the recorded prefix when extraction is malformed, capped, or over budget. ALFWorld extraction may read training labels and native PDDL success conditions, but not hidden initial state or walkthroughs; WebShop excludes hidden targets and ASIN metadata. MiniHack and ScienceWorld use visible tasks and recorded experience without hidden truth. We distinguish malformed outputs, unavailable commands, and executed actions through generation and environment-feedback metadata.

\subsection{GRPO, Optimizers, and Distributed Batching}

We train for 150 RL updates. Each update samples 16 tasks and generates eight independent trajectories per task, yielding 128 base trajectories before localization sampling. Policy rollouts use temperature 1.0. We apply OPD with weight $\lambda_{\rm OPD}=0.01$ only within adaptive teaching intervals. Qwen's native thinking template is disabled while benchmark-required output formats are preserved.

We use action-step GRPO following OCSD, with rewards and interaction budgets set for each environment. We apply the RL objective to the complete recorded trajectory, combining local teaching with task-outcome learning. Reference k3 KL and entropy coefficients are 0.01 and 0.001, respectively; the frozen RL-initial SFT model supplies the reference.

In our current optimizer configuration, we count the eight environment trajectories per task once and set optimizer sampling multiplicity to one. With global agent minibatch $B$ and $W$ ranks, the nominal local batch is $B/W$. Within a logical microbatch, we scale token losses by $D G$, where $D$ is its response-token count and $G$ is the number of gradient-accumulation steps, then average rank gradients. OPD retains the fixed original-decision denominator from Section~\ref{sec:how}.

We optimize all agent parameters using BF16 computation and FP32 AdamW master weights and moments, with online learning rate $10^{-6}$, weight decay 0.01, and gradient norm limit 1.

\subsection{Environment and Generation Budgets}
The step limits for WebShop, ALFWorld, MiniHack, and ScienceWorld are 15, 50, 30, and 30. WebShop and ALFWorld retain the latest two interaction turns; MiniHack and ScienceWorld retain the full history. Response limits are 2,048 tokens for WebShop 4B and MiniHack, and 512 for ALFWorld, ScienceWorld, and the WebShop 1.7B localization comparisons. We disable Qwen's native thinking while preserving each environment's output format.

\Needspace{9\baselineskip}
\section{Supplementary Analyses}

\subsection{Sensitivity to the OPD Weight}
\label{sec:opd-weight-sensitivity}

We vary the OPD weight on ALFWorld seen and WebShop (Table~\ref{tab:opd-weight-sensitivity}). A small weight, 0.01, gives the highest mean on all three metrics. Increasing the weight generally lowers performance, with local fluctuations at intermediate values. This suggests that stronger distillation is not automatically more useful: local guidance needs to be balanced with full-trajectory outcome learning. The same ranking across the two environments motivates the small coefficient used in our training configuration.

\begin{table}[t]
\centering
\caption{OPD-weight sensitivity for Qwen3-1.7B. Success is in percent; WebShop score uses 0--100.}
\label{tab:opd-weight-sensitivity}
\small
\setlength{\tabcolsep}{4pt}
\begin{tabular}{@{}cccc@{}}
\toprule
\multirow{2}{*}{$\lambda_{\rm OPD}$} & ALFWorld & \multicolumn{2}{c}{WebShop} \\
\cmidrule(lr){2-2}\cmidrule(lr){3-4}
& Seen success & Score & Success \\
\midrule
0.01 & $\mathbf{72.1{\pm}1.2}$ & $\mathbf{77.7{\pm}1.8}$ & $\mathbf{65.8{\pm}1.1}$ \\
0.03 & $70.2{\pm}1.3$ & $75.3{\pm}2.3$ & $64.2{\pm}2.3$ \\
0.05 & $69.1{\pm}2.1$ & $72.5{\pm}3.2$ & $64.3{\pm}1.3$ \\
0.08 & $68.3{\pm}0.9$ & $73.8{\pm}3.3$ & $61.3{\pm}2.3$ \\
0.10 & $69.4{\pm}3.1$ & $73.1{\pm}1.2$ & $59.4{\pm}2.2$ \\
\bottomrule
\end{tabular}
\end{table}

\subsection{Protocol for the Introductory Observations}
\label{app:intro-observations}

We use the base Qwen3-1.7B without SFT or RL updates on ALFWorld training tasks. Of 84 screened tasks with eight source rollouts each, we retain 15 having both success and failure and a failed trace of at least ten steps. One failed trace and one successful reference are chosen per task using fixed seeds; all selected failed traces contain 50 actions. We display a seeded-random subset of six curves. For each trace we replay prefixes of 0, 5, \ldots, 50 actions, verify restoration, and run 16 independent continuations. Replayed actions count toward the total 50-action budget; the zero endpoint therefore has no remaining recovery opportunity. Model input retains only the latest two interactions. The selected tasks form a mixed-outcome diagnostic sample, with decreasing continuation budgets at later prefixes.

For panel (b), we score the same recorded replies with identical base weights and response-token prefixes. The teacher receives the complete public successful trajectory of the same task as its privileged context. Its 40,960-token context accommodates the reference, while the actor uses a 2,560-token context. This diagnostic uses the successful-trajectory reference as a common privileged context. We average full-vocabulary JS over action tokens at each turn, sum turn means within ten relative-position bins, and normalize by the trajectory total. All 750 turns are scored. Bins are aligned to the largest adjacent decline estimated from all 16 continuations in panel (a), with out-of-range bins zero-padded; the uniform reference undergoes identical alignment and padding. The plotted band is the across-task interquartile range.

The aligned center contains a mean 19.2\% of JS mass, with task-bootstrap 95\% interval [8.1\%, 32.3\%], compared with a 10\% uniform reference. Selecting a decline interval with the first eight continuations and evaluating it with the other eight gives a mean decline of 5.8 percentage points, with interval [$-8.3$, 20.8]. The split-half interval spans zero, and the uniform reference lies within the JS interval, reflecting variability across the 15 sampled tasks. JS quantifies the change in the action distribution induced by the reference. The 60,771 additional diagnostic actions are separate from branch-free online training.

\Needspace{9\baselineskip}
\section{Benchmarks and Baselines}
\label{app:benchmark-baseline}

\subsection{Benchmarks}
\label{app:benchmarks}

\paragraph{WebShop.}
WebShop is a simulated shopping website built from real product data and natural-language purchase requests~\citep{yao2023webshopscalablerealworldweb}. The agent searches, browses product pages, selects options, and purchases an item satisfying the request. We report mean score on a 0--100 scale and success rate, distinguishing partial requirement satisfaction from full completion.

\paragraph{ALFWorld.}
ALFWorld provides text-based household environments aligned with embodied tasks~\citep{shridhar2021alfworld}. Its six categories require placing, examining, cleaning, heating, cooling, or moving two objects. We report success for each category and episode-weighted Overall success. Seen and unseen splits assess familiar and held-out environment configurations, respectively.

\paragraph{MiniHack.}
MiniHack builds procedurally generated tasks from NetHack's grid worlds and interaction mechanics~\citep{samvelyan2021minihack}. It tests sequences of navigation and object-use decisions under changing observations. We define training tasks by environment seeds and report the fraction of evaluation episodes that reach the task's success condition.

\paragraph{ScienceWorld.}
ScienceWorld tests scientific reasoning through actions in a text-based environment, including inspecting objects, using equipment, and carrying out experiments~\citep{wang2022scienceworld}. We report native task score and binary success separately: progress toward a scientific objective can earn partial score before the whole task is completed.

\subsection{Baselines}
\label{app:baselines}

\noindent\textbf{Vanilla~\citep{qwen2025qwen3} and Skill-prompt}\quad
Vanilla uses the original Qwen3 backbone without task-specific post-training. We use Skill-prompt as an inference-time guidance control: it adds guidance to the context without updating model weights. This control evaluates explicit prompting, related to the trajectory-derived guidance studied by \citet{wang_skill-sd_2026}.

\noindent\textbf{GRPO~\citep{shao2024deepseekmath}}\quad
The reward-only baseline trains on task outcomes using group-relative policy optimization. It provides the comparison for learning from complete-trajectory rewards without an additional distillation objective.

\noindent\textbf{OPSD~\citep{zhao_self-distilled_2026} and GRPO+OPSD~\citep{lu_self-distilled_2026}}\quad
OPSD uses the same model with privileged context as a teacher and distills its token predictions on the student's own responses. The standalone baseline trains through distillation, while GRPO+OPSD directly combines it with the reward-based objective. Together they test whether dense supervision alone, or its direct addition to RL, suffices.

\noindent\textbf{RLSD~\citep{yang_self-distilled_2026}}\quad
RLSD uses differences between the ordinary and privileged-context policies to adjust token-level update magnitudes, while environment rewards determine the update direction. It integrates the distillation signal into reward-guided optimization.

\noindent\textbf{SDAR~\citep{lu_self-distilled_2026}}\quad
SDAR retains RL as its main objective and adds gated self-distillation. Its token-level gate strengthens teacher-endorsed signals and attenuates negative teacher signals, addressing the uneven reliability of privileged guidance in multi-turn interaction.

\noindent\textbf{OCSD~\citep{yang2026agenticreinforcementlearningobservationcalibrated}}\quad
OCSD contrasts matched replay contexts with and without future observations to isolate observation-dependent support. It uses this calibrated signal at uncertain steps to modulate token-level GRPO updates while preserving the outcome-derived update direction.

\end{document}